%% file: main.tex
\documentclass[10pt,twocolumn,letterpaper]{article}

\usepackage{wacv}
\usepackage{times}
\usepackage{epsfig}
\usepackage{graphicx}
\usepackage{amsmath}
\usepackage{amssymb}
\usepackage{booktabs}

\usepackage{multirow}
\usepackage{enumitem}
\usepackage{cite}
\usepackage[table, dvipsnames]{xcolor}
\definecolor{LightBlue}{rgb}{0.86,0.90,0.95}
\definecolor{LightGreen}{rgb}{0.86,0.95,0.90}

\def\wacvPaperID{776} 

\wacvapplicationstrack 

\wacvfinalcopy 

\ifwacvfinal
\usepackage[breaklinks=true,bookmarks=false]{hyperref}
\else
\usepackage[pagebackref=true,breaklinks=true,colorlinks,bookmarks=false]{hyperref}
\fi

\begin{document}
\title{MEVID: Multi-view Extended Videos with Identities \\for Video Person Re-Identification}

\author{Daniel Davila\thanks{Equal Contribution}~~~Dawei Du$^\ast$~~~Bryon Lewis~~~Christopher Funk~~~Joseph Van Pelt\\
Roderic Collins~~~Kellie Corona~~~Matt Brown~~~Scott McCloskey~~~Anthony Hoogs~~~Brian Clipp\\
\vspace{-5pt}\\
\normalsize \textsl{Kitware, NY \& NC, USA}\\
{\tt\small \url{https://github.com/Kitware/MEVID}}
}

\newcommand{\thisdatasetnospace}{MEVID}
\newcommand{\thisdataset}{\thisdatasetnospace~}
\newcommand{\thedataset}{\thisdatasetnospace~}
\newcommand{\comment}[1]{\textcolor{red}{#1}}

\maketitle
\thispagestyle{empty}

\begin{abstract}
\input{abstract}
\end{abstract}

\input{introduction}
\input{related_work}
\input{dataset}
\input{experiment}
\input{conclusion}

{\small
\bibliographystyle{ieee_fullname}
\bibliography{references}
}
\newpage
\appendix
\input{appendix.tex}

\end{document}

%% file: abstract.tex
In this paper, we present the Multi-view Extended Videos with Identities (MEVID) dataset for large-scale, video person re-identification (ReID) in the wild. To our knowledge, MEVID represents the most-varied video person ReID dataset, spanning an extensive indoor and outdoor environment across nine unique dates in a $73$-day window, various camera viewpoints, and entity clothing changes. Specifically, we label the identities of $158$ unique people wearing $598$ outfits taken from $8,092$ tracklets, average length of about $590$ frames, seen in $33$ camera views from the very-large-scale MEVA person activities dataset. 
While other datasets have more unique identities, MEVID emphasizes a richer set of information about each individual, such as: $4$ outfits/identity vs. $2$ outfits/identity in CCVID, $33$ viewpoints across $17$ locations vs. $6$ in $5$ simulated locations for MTA, and $10$ million frames vs. $3$ million for LS-VID. Being based on the MEVA video dataset, we also inherit data that is intentionally demographically balanced to the continental United States. To accelerate the annotation process, we developed a semi-automatic annotation framework and GUI that combines state-of-the-art real-time models for object detection, pose estimation, person ReID, and multi-object tracking.
We evaluate several state-of-the-art methods on MEVID challenge problems and comprehensively quantify their robustness in terms of changes of outfit, scale, and background location. 
Our quantitative analysis on the realistic, unique aspects of MEVID shows that there are significant remaining challenges in video person ReID and indicates important directions for future research. 

%% file: introduction.tex
\section{Introduction}
\label{sec:introduction}
Searching a corpus of video for a person of interest has applications in public safety and security (airports, stadiums, fairs), retail customer-behavior analysis, and in search and rescue. Underpinning the task of automated person search is the sub-task of video-based person re-identification (ReID).
Real-world person ReID is challenging due to complex variations of viewpoint, appearance (such as changes of clothing), person pose, lighting, occlusion, resolution, background, and crowd density settings. 

Various datasets have attempted to provide a strong baseline for solving the person ReID problem, as detailed in Sec.~\ref{sec:related_work}. However, existing works have failed to capture the full set of aforementioned challenges expected during person search deployed on real-world surveillance video.

\begin{figure}[t!]
    \centering
    \includegraphics[width = 0.95\columnwidth ]{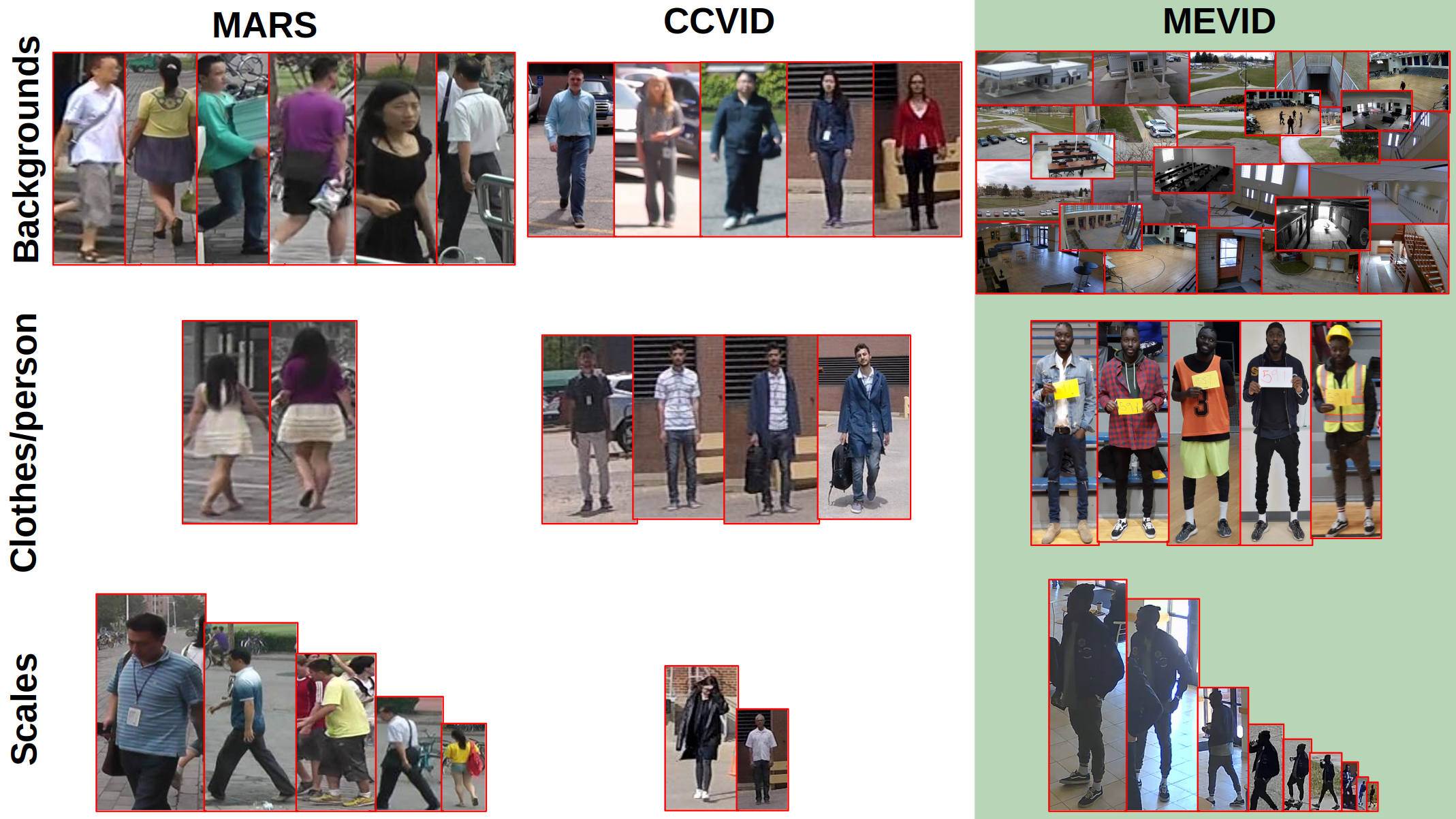}
    \caption{A comparison of \thedataset (highlighted in \textcolor{Green}{green}) to prior person ReID datasets~\cite{DBLP:conf/eccv/ZhengBSWSWT16,ccvid}. Our MEVID dataset significantly increases the diversity of location, viewpoint, number of outfits per person, scales on target, and overall number of tracklets for video person ReID.}
    \label{fig:search_datasets}
    \vspace{-15pt}
\end{figure}

To address this gap, we develop the large-scale Multi-view Extended Videos with Identities (\thisdatasetnospace) dataset. \thedataset provides an extensive and diverse dataset for video person re-identification across a variety of scales, locations, viewpoints, and changes of clothes for each individual target, as shown in Figure~\ref{fig:search_datasets}. We build on top of the MEVA dataset~\cite{DBLP:journals/corr/abs-2012-00914}, which includes hundreds of hours of video in complex indoor/outdoor scenarios, with rich annotations of activities and the tracklets of individuals taking part in those activities. This prior work recorded $176$ actors for three weeks in March and May of 2018, wearing on average $20$ different outfits, performing scripted scenarios and spontaneous background activity. We create the new and challenging MEVID dataset by tracking individuals in these videos to generate person tracklets and then linking these tracklets across viewpoints, locations, days, and outfits throughout the weeks of data collection. MEVA was collected at an access-controlled facility in full compliance with Human Subject Research (HSR) guidelines with all actors signing IRB-approved consent forms; \thedataset inherits this compliance.

Adding global identity information to the activities already annotated in MEVA facilitates the development and evaluation of a number of interesting challenge problems. This includes all of the traditional person ReID problems: image-to-image, image-to-video, and video-to-video. Further, we label each change of clothing for each actor in the dataset, allowing evaluation of change-of-clothing ReID. The change-of-clothing scenario is ubiquitous in practical applications of person search but only supported in the literature by a few limited-scope datasets, see Section~\ref{sec:related_work} for discussion. 
In brief, we split \thedataset into non-overlapping train and test sets. The train set ($6$ dates over $9$ days) contains $104$ global identities with $485$ outfits in $6,338$ tracklets, and the test set ($3$ dates over $5$ days) includes $54$ global identities with $113$ outfits in $1,754$ tracklets.
A detailed comparison of the \thisdataset dataset with other related datasets is presented in Table~\ref{tab:dataset-comparison}. 
Since MEVID is built on top of the MEVA activity dataset, our person labels overlap with the activity labels, and our dataset can be used to identify how many activities are done by the same individual. This will enable detecting compositional, complex activities performed by an individual across different camera views, large spans of time, and long distances. 

This work describes the development of \thedataset and associated challenges problems. We use \thedataset to advance state-of-the-art methods for video-based person ReID, including change-of-clothing ReID, change-of-scale ReID, and change-of-location ReID. Ours is the first paper to evaluate the ability of change-of-clothing ReID across multiple locations, viewpoints, scales, and ambient conditions in the wild.
In our experiments (Sec.~\ref{sec:evaluation}), we discuss and analyze $10$ existing state-of-the-art video person ReID methods~\cite{DBLP:conf/eccv/GuCMZC20, DBLP:conf/aaai/PathakEG20, DBLP:conf/eccv/HouCMSC20, DBLP:journals/tip/WuBLWTZ20, DBLP:conf/iccv/EomLLH21, DBLP:conf/iccv/WangZGGL021, DBLP:conf/cvpr/HouCM0S21, DBLP:journals/corr/abs-2202-06014, ccvid}. Finally, we summarize corresponding issues and future research directions. We hope such comprehensive analysis will boost the research in video person ReID in the wild.
The contributions of this paper are summarized as follows:
\begin{itemize}[align=left,leftmargin=4pt,labelindent=4pt,labelwidth=0pt,labelsep=5pt,topsep=0pt,itemsep=3pt,partopsep=1ex,parsep=0ex]
    \item We \textbf{introduce a video person ReID dataset surpassing all others in data variety and difficulty} by annotating unique person identities associated with tracklets extracted from a large-scale video collection (MEVA).
    \item We \textbf{show key shortcomings of state-of-the-art} video ReID methods in performance on change-of-clothes scenarios, change-of-scale, and change-of-location. 
    \item We \textbf{release the full annotation process and software stack as open-source, including the first open-source tool for video ReID annotation} that was used to develop the dataset to facilitate the further development of person tracking, search, and activity recognition datasets.
\end{itemize}

\begin{table*}[t]
\centering
\setlength{\tabcolsep}{3pt}
\small
\rowcolors{1}{gray!15}{white}
\begin{tabular}{lrrrrrrrrc}
\toprule \rowcolor{LightBlue} 
dataset & \#frames & \#identities & \#outfits & \begin{tabular}[c]{@{}l@{}}\#outfits per\\ identity\end{tabular} & \#BBoxes & \#tracklets & \#viewpoints & \#locations & year \\
\midrule
CUHK-SYSU$^{+}$~\cite{DBLP:conf/cvpr/XiaoLWLW17} & $18,184$ & $8,432$ & $-$ & $-$ & $99,809$ & $-$ & $-$ & $-$ & 2016\\
PRW$^{+}$~\cite{DBLP:conf/cvpr/ZhengZSCYT17} &$11,816$ &$932$ & $-$ & $-$ & $34,304$ & $-$ &$6$ & $1$ &2017\\
\hline
MARS~\cite{DBLP:conf/eccv/ZhengBSWSWT16} & $>1.19$M &$1,261$ &$1,261$& $1$ & $>1.19$M &$20,478$ &$6$ &$-$ &2016\\
iLIDSVID~\cite{DBLP:conf/eccv/LiZG18a} &$43,800$ &$300$ &$300$ &$1$ &$42,460$ &$600$ & $-$ & $-$ & 2018\\
LS-VID~\cite{DBLP:conf/iccv/LiZW0019} &$>2.98$M &$3,772$ &$3,772$ &$1$ &$>2.98$M  &$14,943$ &$15$ & $-$ &2019\\
MTA$^{*}$~\cite{DBLP:conf/cvpr/KohlSSB20} & $25,092$ &$-$ &$2,840$ & $1$ &$>37.3$M & $2,840$ &$6$  & $5$ &2020\\
P-DESTRE~\cite{kumar2020p} &$105,518$ &$253$ & $-$ & $-$ &$>14.8$M & $1,894$ & moving & $2$ &2020\\
CCVID~\cite{ccvid} & $347,833$ & $226$ &$480$&  $\sim2$ & $347,833$ &$2,856 $ &$1$ &$1$ &2022\\
\rowcolor{LightGreen}\thedataset & $>10.46$M & 158 & 598 &  $\sim4$ & $>1.7$M & $8,092$ & $33$ & $17$ &2022 \\
\bottomrule
\end{tabular}
\caption{Comparison between our \thedataset dataset (\textcolor{Green}{green}) and currently released person search datasets, where $^{+}$ denotes image-to-image, and $^{*}$ denotes synthetic datasets. $-$ indicates no data available for this attribute.
}
\label{tab:dataset-comparison}
\vspace{-10pt}
\end{table*}

%% file: related_work.tex
\section{Related Work}
\label{sec:related_work}
{\noindent {\bf Video person ReID datasets.}} 
Our goal with \thedataset is to facilitate the training and evaluation of algorithms for video person ReID in the wild. Several datasets exist in the literature for image-to-image and image-to-video ReID, such as CUHK01-03~\cite{DBLP:conf/cvpr/LiZXW14}, PRW~\cite{DBLP:conf/cvpr/ZhengZSCYT17}, and CSM~\cite{huang2018person}. 
There are also pure video person ReID datasets, including MARS~\cite{DBLP:conf/eccv/ZhengBSWSWT16}, iLIDSVID~\cite{DBLP:conf/eccv/LiZG18a}, LS-VID~\cite{DBLP:conf/iccv/LiZW0019}, P-DESTRE~\cite{kumar2020p}, and synthetic MTA~\cite{DBLP:conf/cvpr/KohlSSB20}. 
However, all of them suffer from a lack of diversity in terms of viewpoints, changes of clothes for each person, scale, and changes in background. 
Recently, Gu~\etal~\cite{ccvid} produced the first video change-of-clothing ReID (CCVID) dataset. This work isolates the impact of clothing changes on the task of video ReID by fixing viewpoint, background, and scale to a relatively limited range while capturing video of actors in multiple outfits. While this demonstrates the shortcomings in the state-of-the-art for CCVID, this work limits the ability to develop models for a wider set of conditions of diversity in locations, actors, viewpoints, both indoor and outdoor scenes, and target scale between query and gallery.

{\noindent {\bf Video person ReID methods.}} 
Different from image person ReID \cite{bot,munjal2019query,DBLP:journals/tmm/LuoJGLLLG20,DBLP:journals/corr/abs-2203-09642} relying on a still image, the goal of video person ReID is to model and match tracklets of the same person across different cameras. 
To extract discriminative representation for noisy tracklets, current methods~\cite{DBLP:conf/cvpr/McLaughlinRM16, DBLP:conf/cvpr/LiB0W18, DBLP:conf/iccv/LiZW0019, DBLP:conf/aaai/PathakEG20, DBLP:conf/iccv/WangZGGL021, DBLP:journals/corr/abs-2202-06014, ccvid} leverage spatio-temporal information for video person ReID.
McLaughlin~\etal~\cite{DBLP:conf/cvpr/McLaughlinRM16} use a simple Siamese network to extract video features based on convolution, recurrent, and temporal pooling layers. 
To remove unavoidable outlier tracking frames, Li~\etal~\cite{DBLP:conf/cvpr/LiB0W18} propose the spatio-temporal attention model to discover a diverse set of distinctive body parts. 
Similarly, by using bag-of-tricks for training, Pathak~\etal~\cite{DBLP:conf/aaai/PathakEG20} add the Attention and CL loss on top of a temporal attention-based neural network to crop out noisy frames.
To adapt to various lengths of tracklets, Li~\etal~\cite{DBLP:conf/iccv/LiZW0019} model both the short-term temporal cues among adjacent frames and the long-term relations among in-consecutive frames.
To deal with occlusions, Wang~\etal~\cite{DBLP:conf/iccv/WangZGGL021} propose the Pyramid Spatial-Temporal Aggregation architecture, which integrates frame-level features hierarchically into a final video-level representation.
In terms of scale variations, Zang~\etal~\cite{DBLP:journals/corr/abs-2202-06014} propose a multi-direction and multi-scale pyramid in Transformer to capture fine-grained part-informed information of people.
Recently, to deal with change-of-clothing problem in real world, Gu~\etal~\cite{ccvid} develop a simple baseline method by using Clothes-based Adversarial Loss, which mines clothes-irrelevant features from the original RGB images.

{\noindent {\bf Annotation tools for video person ReID.}} 
To our knowledge, there do not exist any free and open-source video ReID annotation tools for use in the community. Several tools exist such as CVAT~\cite{cvat} and LabelMe~\cite{labelme} which provide basic functionality in terms of bounding box annotation, trajectory annotation, automated annotation bootstrapping, and even attribute labeling, No tool directly facilitates inter-track linking and build-up of global entity information for all unique persons or objects that appear in a dataset. Without this ability, linking instances of persons across videos is tedious and time-consuming. 

%% file: dataset.tex
\section{\thedataset Dataset}
\label{sec:dataset}
Our goal in \thedataset is to create an extensive set of person tracklets from the MEVA dataset~\cite{DBLP:journals/corr/abs-2012-00914} and annotate the person's global identity for each tracklet. These identity-annotated tracklets enable our development and evaluation of video person ReID methods in the wild with real-world challenges such as change-of-clothing, large-scale variation, multiple locations, and more.

\subsection{Data Collection}
The video used for our \thedataset dataset is a subset of the roughly $9,300$ hours of video collected for the MEVA dataset. MEVA contains $37$ types of activity tracklets for $176$ actors across $144$ hours of video, totaling $66,172$ annotated activities. 
We add additional tracklets to the dataset and link actors across video views and outfits in order to create the most diverse video ReID dataset to date. 
The new \thedataset dataset contains hundreds of hours of videos, over $100$ actors in multiple sets of clothing at an access-controlled facility with indoor and outdoor scenes for nine unique dates across a $73$-day window. 
Notably, the videos are taken from over $30$ ground-level cameras both inside and outside, capturing people at a wide range of distances, angles, and lightning conditions.

\textbf{Ethical Considerations.}
The baseline MEVA dataset, from which this work is derived, was collected under rigorous oversight of an independent Institutional Review Board (IRB). MEVA contains only hired actors and persons who consented to being collected on video, filmed in a highly controlled environment rented for the data collection. 
In \thedataset dataset, the person identities or attributes are originally authorized for the MEVA video collection. All persons under observation are consenting persons. 
None of the events, places or actions have significance in the real world as the entire collection was staged and subject identities are not known. The data is collected to achieve intentional demographic alignment of the dataset with the continental united states, in terms of racial and gender identity. 

\subsection{Dataset Annotation}
\label{sec:dataset_annotation}
\begin{figure}[t]
    \centering
    \includegraphics[width=0.95\columnwidth ]{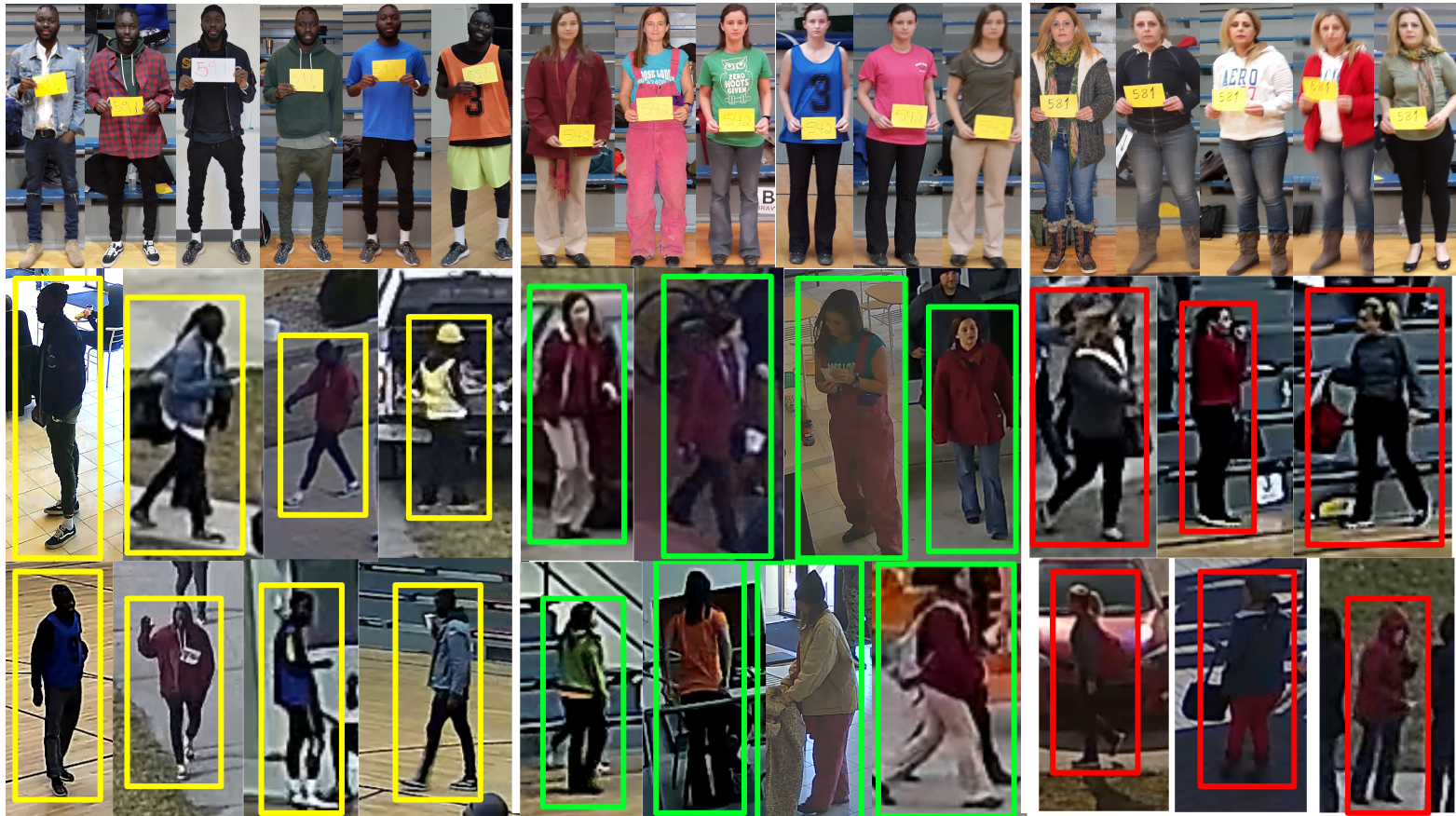}
    \caption{Examples of the actor checkin photos (top row), which are used to associate tracklets in the \thedataset video (middle and bottom rows) with a global outfit and identity ID.}
    \label{fig:checkin_photos}
    \vspace{-10pt}
\end{figure}

The MEVA dataset contains over $2,237$ unique outfits, worn by $176$ actors for three weeks over a two month collection window. A very small subset of this video data was labelled prior to this work, and these labels are purely tracklets with no ID information. During data collection, each actor was photographed, front and back, with and without outerwear. Examples of these checkin photos from the dataset are shown in Figure~\ref{fig:checkin_photos}. Each photo shows the actor holding a card which identifies their GPS logger ID. The method by which we annotate the MEVA video dataset with global IDs is to link each instance of a person in the videos with a particular checkin photo, which itself is assigned metadata that encodes both the outfit ID (\eg, pants and shirt) and global identity (\eg, Alice or Bob). 

We propose two tiers of video ReID problem. The first is the commonly studied problem of same-clothing person ReID, which is the task of re-associating across time or viewpoint two instances of the same person in the same outfit. The second is that of change-of-clothing ReID, which is the task of re-associating instances of the same person across time or viewpoints after a change of clothes has taken place. We also capture both same-clothing and change-of-clothing examples of persons across many different scales, locations, angles, and lighting conditions. 

To improve the efficiency of labeling large-scale video person ReID datasets, we have developed the free, open source, and semi-automatic \thedataset annotation framework and GUI. First, we extract tracklets from video data by running a combination of state-of-the-art real-time models for object detection, pose estimation, person ReID, and multi-object tracking. Second, we ignore the tracklets with low detection confidences or small scales. Third, we conduct a fast linking of tracklets to checkin photos subject to constraints as discussed below. 

The tracker, raw tracker output on all \thedataset videos, and annotation GUI are freely available for future teams to append this level of ground truth to the dataset. The GUI facilitates these steps with a web-based interface monitoring the progress of each video through the process. These stages, discussed below, include tracklet generation, tracklet ingest into the GUI, tracklet cleanup, tracklet review, and tracklet linking to nexus chips which serve as anchor for each global identity, in this case we use the MEVA actor checkin photos. The GUI can be readily provisioned on AWS which allows rapid global collaboration and acceleration of tracking and other support workflows for annotation. Within the interface, we implement tools for bounding box and tracklet annotation and cleanup, including many tricks and features derived from hundreds of hours of annotation time.

\subsubsection{Extracting Tracklets from Video Data}
\label{sec:dataset_tracking}
\begin{figure}[t]
    \centering
    \includegraphics[width=0.9\columnwidth ]{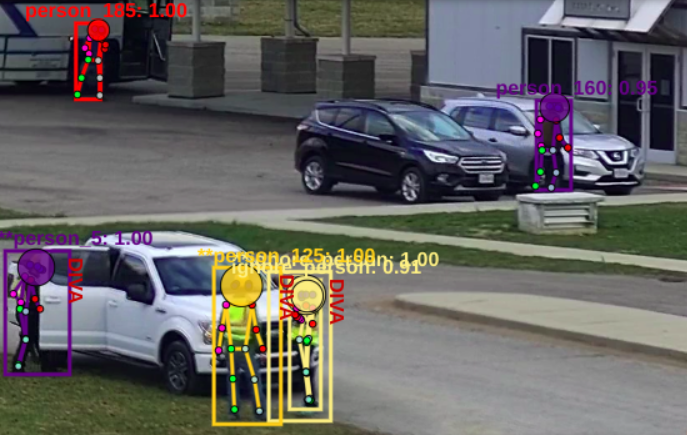}
    \caption{Examples of \thedataset tracklet annotations. We use a head tracking point first to disambiguate persons from one another, then full body pose when head features are ambiguous.}
    \label{fig:annotation_examples}
    \vspace{-10pt}
\end{figure}

Due to the ambiguity of bounding boxes generated by the CenterNet2 detector~\cite{centernet2} in crowded scenes, we use HRNet~\cite{hrnet} to exploit pose information in the box as the canonical track state. Then, the FairMOT tracker~\cite{DBLP:journals/ijcv/ZhangWWZL21} is used to calculate tracklets.
As shown in Figure~\ref{fig:annotation_examples}, we present a tiered approach to keypoint representation between head centroid and full body, which accelerates the annotation task by reducing the total signal an annotator must consider when verifying that a tracklet is free of switches and breaks.
Note that poses are not corrected by the annotation process unless there is ambiguity as to which person the pose refers to, in which case certain joints are moved in order to eliminate that ambiguity. For this reason, \thedataset is not a pose tracking or pose ReID dataset, but rather the poses are used as guides during tracklet-to-checkin photo linking for the annotator, and later for model developers, to understand which person is assigned a particular global ID. 

\subsubsection{Ignored Tracklets During Annotation}
Some tracklets are ignored to speed up the annotation process later. 
First, we assign an ignore label to all detections smaller than a person size threshold to keep the minimum resolution on target required to visually ReID tracklets to their checkin photos at annotation time. In this work, we ignore detections with less than an empirically-derived height of $75$ pixels or width of $25$ pixels.
Second, we ignore any detections which overlap more than $0.3$ IoU with one another in the set. Due to the nature of current state-of-the-art in tracking, high density crowd scenes cause ID switches, significantly delaying the cleanup of the person tracklets, which require the pose-to-boxes and poses across all track states to unambiguously refer to one person. 
Across the videos being cleaned, we require that only a single track state is associated with a single instance of a person on any given frame, and the tracklets are clean with no switches. 

\begin{figure}[t]
\centering
\includegraphics[width=0.9\linewidth]{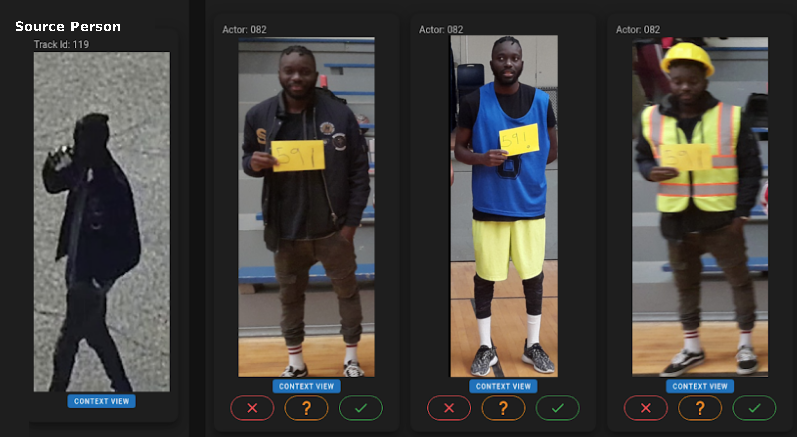}
\caption{Screenshot of the Linker user interface. On the left a subject detection from the video is shown and then compared to probable checkin photos on the right. Annotators can select ``Yes'', ``No'', or ``Don't Know''.} 
\label{fig:linker}
\end{figure}

\subsubsection{Tracklet Cleanup and Linking to Global IDs}
After filtering out all the ignored tracklets, a second annotator performs an audit, \ie, elevating the video tracklets to the linking stage. We exploit a myriad of constraints to facilitate rapid linking of tracklets to checkin photos. 

The first constraint is to compute a correspondence between any available GPS tracklets (if available) and the video space tracklets from our tracker. By projecting our tracklets into world space, we can narrow down the list of potential matches between a tracklet and checkin photo by checking which GPS trackers reported a location in the vicinity of the tracklet and cross-referencing this list of GPS IDs with the associated checkin photos, which are each tagged with a GPS ID. 

Then, given the specific date which actor outfits appeared on, we can reduce the search space for the annotator by over 100x times when trying to match a tracklet to one of over $4000$ checkin photos. 
For each tracklet being linked, the annotator is presented with a filtered and sorted list of potential checkin photos based on the distance between the with the Bag of Tricks ReID descriptor~\cite{DBLP:journals/ijcv/ZhangWWZL21} computed for each track state and the checkin photos. 
Within the linker, annotators are enabled to choose one of three options (Yes, No, or Don't Know) when determining whether a tracklet and checkin photo represent the same person, as shown in Figure~\ref{fig:linker}.
We record each of these decisions in a tracklet-to-checkin photo affinity matrix, which is used in combination with the spatio-temporal filtering described above to make a graph cut of the matching space and reduce the possible number of comparisons from tracklets to checkin photos. 

\subsection{Dataset Statistics}
\begin{figure*}[t]
\centering
\includegraphics[width=0.9\linewidth]{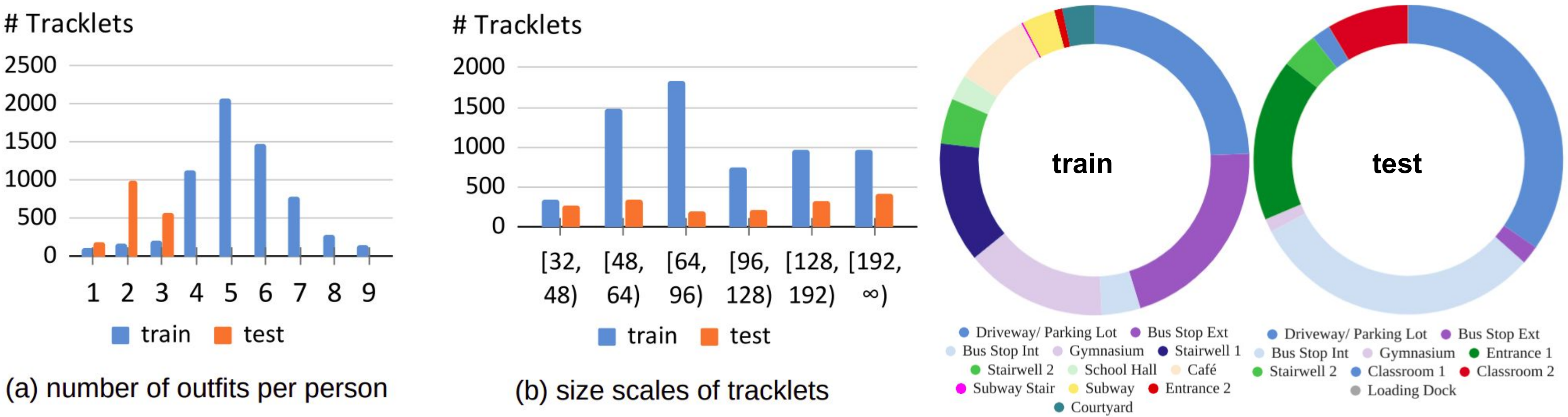}
\caption{Data distribution of outfits, locations and target scales present in \thisdatasetnospace.} 
\label{fig:target_scales}
\end{figure*}

As shown in Figure \ref{fig:target_scales}, \thedataset is divided into the train and
test sets. The train set, $6$ dates over a $9$-day window, contains $104$ global identities with $485$ outfits in $6,338$ tracklets. The test set, $3$ dates over a $5$-day window, includes $54$ global identities with $113$ outfits in $1,754$ tracklets. The test and train sets span a $73$-day window. The frame length per tracklet varies from $1$ to $1,000$, averaging $592.6$ frames.
In terms of the test set, we select $316$ query tracklets such that each query with the specific global identity and outfit combination has correct matches in the remaining $1,438$ gallery tracklets. In addition, query and gallery are captured by different cameras. Each identity will have at least one query under each camera.

Notably, we split camera viewpoints between train and test sets such that we do not bias the person ReID results by testing on the same backgrounds as are present in the train set. Similarly, we seek to prevent the training from over-fitting to actors present in the train set. For this reason, we also split the set of actors into training actors and test actors. This is possible due to the fact that different actors were present on different days of the dataset collection. With the information of who was present on which days, we can construct training sets from the training camera views containing the train actor subgraph of persons, and test sets from the respective camera views and actor buckets for those splits.

\subsection{Relation with Existing Datasets} \label{sec:dataset_statistics}
As presented in Table~\ref{tab:dataset-comparison}, our proposed dataset differs from existing datasets in the following key aspects.
\begin{itemize}[align=left,leftmargin=4pt,labelindent=4pt,labelwidth=0pt,labelsep=5pt,topsep=0pt,itemsep=3pt,partopsep=1ex,parsep=0ex]
\item {\bf 3D models and registered cameras.} Inherited from MEVA~\cite{DBLP:journals/corr/abs-2012-00914}, our dataset contains a 3D model of the entire outdoor site and each exterior camera is registered to this model. This enables site-aware methods that reason about multiple cameras at a single facility to perform tracking, learn typical transit patterns across cameras (camera network topology), and much more. These are unique among any of the other real-world ReID datasets.

\item {\bf Clothes change.} We seek to push the state-of-the-art in video person ReID in the wild, which is currently limited to re-association of persons from different views wearing nearly the same outfits. This is driven by a lack of sufficient one-to-many person-to-outfit datasets. Our dataset contains $158$ actors wearing $598$ different outfits, \ie, $1\sim9$ outfit changes for each actor. This will facilitate the training of novel video search and ReID methods that generalize beyond applications of matching like-outfits. 

\item {\bf Diverse locations.} We collect the video data in both indoor and outdoor scenes. Of the person ReID datasets we have studied, only MSMT17~\cite{DBLP:conf/cvpr/WeiZ0018} includes both indoor and outdoor scenes. All end-to-end person search datasets have one of either indoor or outdoor scenes. In contrast, across $17$ unique locations in \thedataset, we have $33$ different camera viewpoints in total. The videos are collected from a wide range of natural scenes such as parking lots, bus stations, cafes, school environments, and more. 
Additionally, the venues which make up \thedataset are connected spatially via doorways, stairwells, tunnels, and hallways which present interesting transition points and viewpoint changes between exiting one scene and entering another. We release 3D models of all exterior scenes allowing for higher-order inferencing and constraining of the person ReID problem via consideration of spatio-temporal reasonableness of tracklets computed, or in the kinematic plausibility of locomotion throughout the scene.  However, other datasets either have to render to have multiple locations such as MTA~\cite{DBLP:conf/cvpr/KohlSSB20}.  

\item {\bf Target Scales} Of importance to person ReID is a robustness to changes in scale on a person from query to gallery. Our dataset presents a wide range of scales on each target, with the smallest on the order of $75$ pixels and the largest of $500$ pixels on target height. This reflects the difference in scale of the locations being observed around the site of data collection, some of which are intimate indoor settings and others sweeping outdoor scenes of driveways and parking lots in between the buildings of the site. This scale variation will allow us to evaluate the performance of cross-scale ReID, and the impact of significant differences in target resolution on feature embedding. Also present in our dataset is a wide variety of target shapes and aspect ratios. We have many tracklets of persons sitting, or occluded from the waist down, which present a unique challenge when processing chip sequences.
\end{itemize}

%% file: experiment.tex
\section{Baseline Performance} 
\label{sec:evaluation}
We conducted comprehensive experiments on \thedataset to establish a performance baseline. Specifically, we compare $10$ state-of-the-art video person ReID algorithms, including
AP3D~\cite{DBLP:conf/eccv/GuCMZC20}, Attn-CL~\cite{DBLP:conf/aaai/PathakEG20}, Attn-CL+rerank~\cite{DBLP:conf/aaai/PathakEG20}, TCLNet~\cite{DBLP:conf/eccv/HouCMSC20},
AGRL~\cite{DBLP:journals/tip/WuBLWTZ20}, STMN~\cite{DBLP:conf/iccv/EomLLH21}, PSTA~\cite{DBLP:conf/iccv/WangZGGL021}, BiCnet-TKS~\cite{DBLP:conf/cvpr/HouCM0S21}, 
PiT~\cite{DBLP:journals/corr/abs-2202-06014}, and CAL~\cite{ccvid}. We follow the default settings in each paper. Note that CAL~\cite{ccvid} is the only one that explicitly compensates for clothing changes.

\subsection{Experimentation Protocol} \label{sec:video_person_sesarch_evaluation}
Similar to the MARS protocol~\cite{DBLP:conf/eccv/ZhengBSWSWT16, DBLP:conf/iccv/ZhengSTWWT15}, we use mean Average Precision (mAP) and Cumulative Matching Characteristic (CMC) metrics to evaluate ReID methods. We calculate the similarity scores between query and gallery tracklets, then rank the gallery based on these scores. The CMC metric indicates the true match being found within the first $k$ ranks ($k=1,5,10,20$), and the mAP metric focuses on precision and recall over all the queries.

As discussed above, the detailed annotation metadata of \thisdataset allows us to evaluate the performance of algorithms under a variety of challenging factors of increasing difficulty: 
1) \textbf{Change-of-clothing} attribute indicates different outfits for the same person. In the test set, there are $1\sim3$ outfits for each person.
2) \textbf{Location difference} attribute indicates that the tracklet was recorded in indoor or outdoor scenes. The test data contains $11$ cameras in various places including \textit{Driveway/Parking Lot}, \textit{Bus Stop}, \textit{Classroom}, \textit{Entrance} and \textit{Stairwell}. 
3) \textbf{Scale variation} attribute indicates the average person size in the tracklet. Following the COCO protocol~\cite{DBLP:conf/eccv/LinMBHPRDZ14}, two scale categories  are found in our dataset, \textit{Medium} ($[32,96)$) and \textit{Large} ($[96,\infty)$).

\begin{table}[t]
\centering
\rowcolors{1}{gray!15}{white}
\resizebox{\columnwidth}{!}{
\begin{tabular}{lccccc}
\toprule \rowcolor{LightBlue}
\bf{method} & \bf{mAP} & \bf{top-1} & \bf{top-5} & \bf{top-10} & \bf{top-20}\\
 \cmidrule(r){1-1} \cmidrule(r){2-6}
CAL~\cite{ccvid} &\textbf{27.1\%}& \textbf{52.5\%} & \textbf{66.5\%} & \textbf{73.7\%} &\textbf{80.7\%}\\
AGRL~\cite{DBLP:journals/tip/WuBLWTZ20} &19.1\% &48.4\% &62.7\% &70.6\% &77.9\%\\
BiCnet-TKS~\cite{DBLP:conf/cvpr/HouCM0S21} &6.3\% &19.0\% &35.1\% &40.5\% &52.9\%\\
TCLNet~\cite{DBLP:conf/eccv/HouCMSC20} &23.0\% &48.1\% &60.1\% &69.0\% &76.3\%\\
PSTA~\cite{DBLP:conf/iccv/WangZGGL021} &21.2\% &46.2\% &60.8\% &69.6\% &77.8\\
PiT~\cite{DBLP:journals/corr/abs-2202-06014} &13.6\% &34.2\% &55.4\% &63.3\% &70.6\%\\
STMN~\cite{DBLP:conf/iccv/EomLLH21} &11.3\%	&31.0\%	&54.4\%	&65.5\%	&72.5\%\\
Attn-CL~\cite{DBLP:conf/aaai/PathakEG20} &18.6\% &42.1\% &56.0\% &63.6\% &73.1\%\\
Attn-CL+rerank~\cite{DBLP:conf/aaai/PathakEG20} &25.9\% &46.5\% &59.8\% &64.6\% &71.8\%\\
AP3D~\cite{DBLP:conf/eccv/GuCMZC20} &15.9\% &39.6\%	&56.0\%	&63.3\%	&76.3\%\\
\bottomrule
\end{tabular}
}
\caption{Comparison of state-of-the-art video person ReID methods on MEVID. Highest scores are in \textbf{bold}.}
\label{tab:overall-results}
\vspace{-15pt}
\end{table}

\subsection{Results Analysis} 
\label{sec:overall_results}
Table \ref{tab:overall-results} shows the performance of current video person ReID methods on \thisdatasetnospace.
Compared with performance on previous video person ReID datasets, these algorithms perform much worse on our dataset even though the number of individuals is smaller. The top mAP scores on other datasets are $88.5\%$ for MARS~\cite{DBLP:conf/eccv/ZhengBSWSWT16}, $92.1\%$ for iLIDSVID~\cite{DBLP:conf/eccv/LiZG18a}, $69.2\%$ for LS-VID~\cite{DBLP:conf/iccv/LiZW0019}, $80.6\%$ for P-DESTRE~\cite{kumar2020p}, and $81.3\%$ for CCVID~\cite{ccvid}. CAL~\cite{ccvid} introduces a clothes classifier on top of the ReID model to decouple clothing-independent information, achieving the best \thedataset performance of $27.1\%$ mAP and $52.5\%$ top-1 scores.
\thedataset provides more realistic variations in clothes, locations and scales, making it much more difficult for existing algorithms to model discriminative representations for each individual, resulting in inferior performance in the wild. 
In the following, we analyze how the above methods perform in terms of each attribute in detail.

\subsubsection{Change-of-clothing Challenge} \label{sec:clothes_results}
While previous work yields good results in limited diversity datasets, our benchmark presents a unique challenge, the change-of-clothing problem in real-world scenarios. Two test settings are defined to calculate accuracy for an individual: 1) all tracklets have the same clothes as the query; and 2) all tracklets have different clothes from the query.

As shown in Table~\ref{tab:clothes-results}, all existing methods perform poorly on real-world change-of-clothing scenarios. 
For example, the best accuracy from AGRL~\cite{DBLP:journals/tip/WuBLWTZ20} and PSTA~\cite{DBLP:conf/iccv/WangZGGL021} is $\sim5\%$, demonstrating the failure of state-of-the-art video person ReID in complex scenes.
On the other hand, video person ReID methods perform better for the classical video ReID problem of the same outfit. However, the results are still not satisfactory, \ie, $39.0\%$ mAP and $56.6\%$ top-1 scores by the best, CAL~\cite{ccvid}. This is because of location difference and scale variations in the dataset.

\begin{table}[t]
    \centering
    \rowcolors{1}{gray!15}{white}
    \resizebox{\columnwidth}{!}{
    \begin{tabular}{lrrrrrrrrrr}
    \toprule \rowcolor{LightBlue}
    \textbf{Method} & \multicolumn{5}{c}{\textbf{Same Clothes (\%)}} & \multicolumn{5}{c}{\textbf{Different Clothes (\%)}} \\ \cmidrule(r){1-1} \cmidrule(r){2-6} \cmidrule(l){7-11}
          & mAP   & top-1 & top-5 & top-10 & top-20 & mAP   & top-1 & top-5 & top-10 & top-20 \\
    CAL~\cite{ccvid}   & \textbf{39.0}  & \textbf{56.6}  & \textbf{70.8}  & \textbf{78.1}  & \textbf{85.4}  & 4.3   & 3.5   & 10.6  & 14.8  & 19.4 \\
    AGRL~\cite{DBLP:journals/tip/WuBLWTZ20}  & 32.6  & 51.4  & 64.9  & 73.6  & 80.9  & \textbf{5.7}   & 4.9   & \textbf{15.1}  & 19.0  & 25.7 \\
    BiCnet-TKS~\cite{DBLP:conf/cvpr/HouCM0S21} & 8.0   & 20.5  & 36.5  & 41.7  & 51.4  & 1.7   & 0.7   & 4.6   & 7.8   & 13.4 \\
    TCLNet~\cite{DBLP:conf/eccv/HouCMSC20} & 31.9  & 51.7  & 63.5  & 71.9  & 79.2  & 3.9   & 3.5   & 8.8   & 14.1  & 21.1 \\
    PSTA~\cite{DBLP:conf/iccv/WangZGGL021}  & 29.7  & 49.0  & 63.9  & 72.2  & 78.5  & 5.1   & \textbf{5.6}   & 12.3  & \textbf{19.4}  & \textbf{28.9} \\
    PiT~\cite{DBLP:journals/corr/abs-2202-06014}   & 19.5  & 36.8  & 58.7  & 66.3  & 73.6  & 2.0   & 1.1   & 5.3   & 8.5   & 13.7 \\
    STMN~\cite{DBLP:conf/iccv/EomLLH21}  & 18.5  & 33.7  & 58.3  & 69.1  & 76.4  & 1.2   & 0.4   & 1.8   & 3.9   & 6.0 \\
    Attn-CL~\cite{DBLP:conf/aaai/PathakEG20} & 24.2  & 44.4  & 59.7  & 66.3  & 72.6  & 3.4   & 2.8   & 8.5   & 15.5  & 24.6 \\
    Attn-CL+rerank~\cite{DBLP:conf/aaai/PathakEG20} & 34.1  & 50.7  & 63.2  & 68.1  & 72.9  & 4.2   & 2.1   & 9.2   & 13.7  & 22.5 \\
    AP3D~\cite{DBLP:conf/eccv/GuCMZC20}  & 23.2  & 42.7  & 59.7  & 67.7  & 79.2  & 2.9   & 1.8   & 7.4   & 9.5   & 16.6 \\
    \bottomrule
    \end{tabular}
    }
    \caption{Comparison of video person ReID methods in terms of change-of-clothing attribute. Highest scores are in \textbf{bold}.}
    \label{tab:clothes-results}
    \vspace{-4pt}
\end{table}

\begin{figure}[t]
     \centering
     \includegraphics[ width = 1.0\columnwidth ]{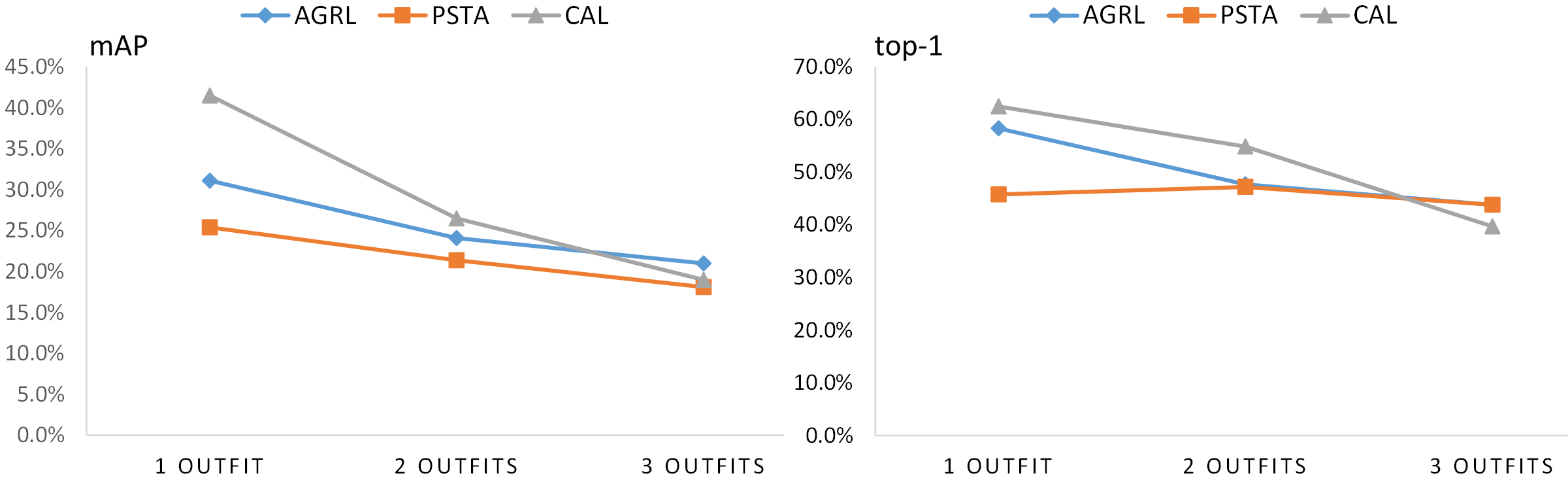}
     \caption{Influence of different number of outfits per person.}
     \label{fig:fine-clothes-results}
     \vspace{-8pt}
\end{figure}

\textbf{Influence of the number of outfits.} We explore the performance of top $3$ methods for different clothes in terms of different number of outfits ($1\sim3$). As shown in Figure \ref{fig:fine-clothes-results}, it can be seen that both mAP and top-1 scores are reduced along with the increasing number of outfits per person. Intuitively, more outfits bring large intra-class variances, which is the core problem in ReID problem in the wild.

\subsubsection{Location Difference Challenge} \label{sec:location_results}
In order to assess the impact of changing background location on the performance of video person ReID, we set up two test settings with the same experimental setup as previously described when testing same- and change-of-clothing ReID, except now we enforce that query and gallery tracklets must have different locations, \ie, indoors/outdoors scenes. 
The results in Table~\ref{tab:locations-results} indicate that there is no substantial impact on the location for mAP score. However, the top-1 score is significantly affected from indoor/outdoor to outdoor/indoor scenes. This is due to domain shift in different locations.
We notice that AGRL~\cite{DBLP:journals/tip/WuBLWTZ20} obtains better results in terms of different locations. 
It employs a structure-aware spatio-temporal graph representation based on pose alignment and feature affinity connections. 
In this way, contextual interaction between relevant regional features can help distinguish people at different places.

\begin{table}[t]
  \centering
    \rowcolors{1}{gray!15}{white}
    \resizebox{\columnwidth}{!}{lk
    \begin{tabular}{lrrrrrrrrrr}
    \toprule \rowcolor{LightBlue}
    \textbf{Method} & \multicolumn{5}{c}{\textbf{Same Locations (\%)}} & \multicolumn{5}{c}{\textbf{Different Locations (\%)}} \\ \cmidrule(r){1-1} \cmidrule(r){2-6} \cmidrule(l){7-11}
          & mAP   & top-1 & top-5 & top-10 & top-20 & mAP   & top-1 & top-5 & top-10 & top-20 \\
    CAL~\cite{ccvid}   & \textbf{24.7}  & \textbf{42.1}  & \textbf{56.6}  & 63.2  & \textbf{72.0} & 22.2  & 35.0  & 49.8  & 55.2  & 62.0   \\
    AGRL~\cite{DBLP:journals/tip/WuBLWTZ20}  & 18.1  & 27.6  & 42.8  & 48.8  & 57.6 & \textbf{22.5}  & \bf{41.1}  & \bf{57.6}  & \bf{64.8}  & \bf{70.1}  \\
    BiCnet-TKS~\cite{DBLP:conf/cvpr/HouCM0S21} & 5.1   & 14.5  & 25.3  & 30.6  & 37.5 & 4.7   & 9.4   & 19.5  & 24.6  & 34.3   \\
    TCLNet~\cite{DBLP:conf/eccv/HouCMSC20} & 20.7  & 38.8  & 52.6  & 60.9  & 68.8 & 18.8  & 33.0  & 42.4  & 48.5  & 55.9   \\
    PSTA~\cite{DBLP:conf/iccv/WangZGGL021}  & 20.0  & 36.8  & 54.6  & \textbf{63.5}  & 71.4 & 16.5  & 28.6  & 41.4  & 49.5  & 57.6   \\
    PiT~\cite{DBLP:journals/corr/abs-2202-06014}   & 12.1  & 25.0  & 45.4  & 53.6  & 60.5 & 10.1  & 19.2  & 34.0  & 39.1  & 49.2   \\
    STMN~\cite{DBLP:conf/iccv/EomLLH21}  & 10.1  & 16.8  & 31.0  & 36.1  & 43.1 & 10.0  & 22.2  & 41.6  & 52.1  & 58.4   \\
    Attn-CL~\cite{DBLP:conf/aaai/PathakEG20} & 16.1  & 35.1  & 48.2  & 54.6  & 64.9 & 14.2  & 26.4  & 40.7  & 47.8  & 55.6   \\
    Attn-CL+rerank~\cite{DBLP:conf/aaai/PathakEG20} & 23.9  & 41.5  & 53.4  & 58.1  & 63.9 & 19.6  & 33.9  & 44.7  & 50.2  & 55.9   \\ 
    AP3D~\cite{DBLP:conf/eccv/GuCMZC20}  & 14.5  & 31.0  & 45.1  & 51.8  & 63.3 & 12.0  & 24.1  & 37.6  & 43.4  & 50.9   \\
    \bottomrule
    \end{tabular}
    }
    \caption{Comparison of video person ReID methods in terms of location difference attribute. Highest scores are in \textbf{bold}.}
    \label{tab:locations-results}
\end{table}

\begin{figure}[t]
     \centering
     \includegraphics[width = 1.0\columnwidth ]{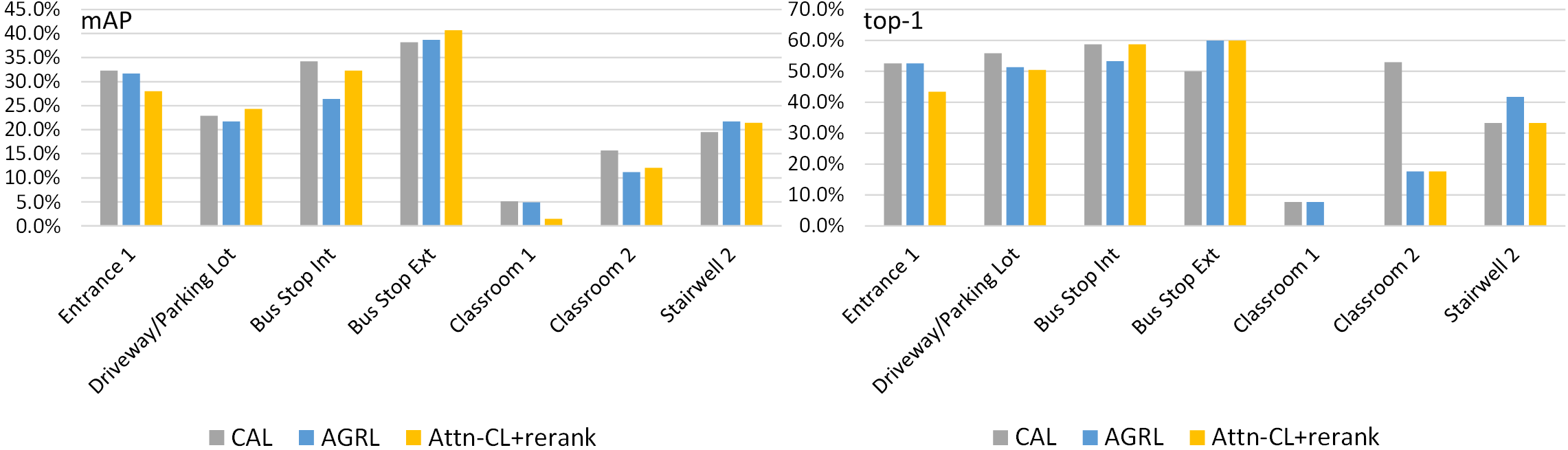}
     \caption{Comparison of performance at specific locations in indoor/outdoor scenes. For each experiment, the query location is fixed but the gallery can take any location in the dataset.}
     \label{fig:fine-location-results}
\end{figure}

\textbf{Comparison of performance at specific locations.} To explore the performance of existing methods w.r.t. locations, we break down indoor/outdoor scenarios into $7$ specific locations. 
As shown in Figure \ref{fig:fine-location-results}, we can conclude that the ReID performance of all the selected algorithms on indoor scenes (\textit{Classroom} and \textit{Stairwell}) is much worse than that in outdoor scenes (\textit{Entrance}, \textit{Driveway/Parking Lot}, and \textit{Bus Stop}). This is maybe because more occlusions in indoor scenes bring noise in feature learning.

\subsubsection{Scale Variation Challenge} \label{sec:scale_results}
Based on the setting in Sec. \ref{sec:video_person_sesarch_evaluation}, we explore the influence of scale variation on video person ReID. 
From Table~\ref{tab:scales-results}, we observe a similar trend as in the location difference attribute evaluation.
Concretely, both Attn-CL+rerank~\cite{DBLP:conf/aaai/PathakEG20} and CAL~\cite{ccvid} perform the best among all the compared methods. 
It is worth noting that Attn-CL+rerank~\cite{DBLP:conf/aaai/PathakEG20} boosts the ReID performance based on the baseline Attn-CL method by using the rerank post-processing \cite{DBLP:conf/aaai/FuWWH19}.

\begin{table}[t]
  \centering
    \rowcolors{1}{gray!15}{white}
    \resizebox{\columnwidth}{!}{
    \begin{tabular}{lcccccccccc}
    \toprule \rowcolor{LightBlue}
    \textbf{Method}& \multicolumn{5}{c}{\textbf{Same Scales (\%)}} & \multicolumn{5}{c}{\textbf{Different Scales (\%)}} \\ \cmidrule(r){1-1} \cmidrule(r){2-6} \cmidrule(l){7-11}
          & mAP   & top-1 & top-5 & top-10 & top-20 & mAP   & top-1 & top-5 & top-10 & top-20 \\
    CAL~\cite{ccvid}   & \bf{24.3}  & \bf{42.3}  & 56.3  & 61.3  & \bf{71.7}  & 20.6  & \bf{35.2}  & \bf{50.7}  & \bf{58.4}  & \bf{62.4} \\
    AGRL~\cite{DBLP:journals/tip/WuBLWTZ20}  & 22.1  & 40.3  & \bf{57.3}  & 64.7  & 71.0  & 17.7  & 29.5  & 43.6  & 49.3  & 58.4 \\
    BiCnet-TKS~\cite{DBLP:conf/cvpr/HouCM0S21} & 5.2   & 14.7  & 26.0  & 31.0  & 39.7  & 4.6   & 10.7  & 20.8  & 25.5  & 34.9 \\
    TCLNet~\cite{DBLP:conf/eccv/HouCMSC20} & 20.7  & 40.0  & 52.3  & 61.0  & 65.0  & 17.6  & 34.2  & 44.3  & 50.3  & 59.4 \\
    PSTA~\cite{DBLP:conf/iccv/WangZGGL021}  & 18.6  & 34.3  & 51.3  & 60.3  & 67.0  & 16.8  & 29.9  & 44.3  & 51.7  & 60.1 \\
    PiT~\cite{DBLP:journals/corr/abs-2202-06014}   & 11.4  & 23.7  & 44.0  & 53.3  & 60.3  & 10.6  & 23.5  & 37.3  & 41.6  & 51.0 \\
    STMN~\cite{DBLP:conf/iccv/EomLLH21}  & 10.5  & 19.2  & 33.2  & 39.2  & 46.2  & 9.4   & 22.3  & 42.0  & 51.0  & 58.0 \\
    Attn-CL~\cite{DBLP:conf/aaai/PathakEG20} & 15.4  & 31.3  & 50.0  & 56.0  & 64.3  & 14.3  & 26.5  & 37.6  & 46.3  & 55.7 \\
    Attn-CL+rerank~\cite{DBLP:conf/aaai/PathakEG20} & 23.1  & 35.7  & 54.0  & 59.3  & 67.7  & \bf{21.2}  & 33.6  & 44.0  & 50.0  & 54.4 \\
    AP3D~\cite{DBLP:conf/eccv/GuCMZC20}  & 14.2  & 31.0  & 47.3  & 53.0  & 63.7  & 11.4  & 24.5  & 35.9  & 42.6  & 52.7 \\
    \bottomrule
    \end{tabular}
    }
    \caption{Comparison of video person ReID methods for scale variation attribute. Highest scores are in \textbf{bold}.}
    \label{tab:scales-results}
\end{table}

\begin{figure}[t]
     \centering
     \includegraphics[ width = 1.0\columnwidth ]{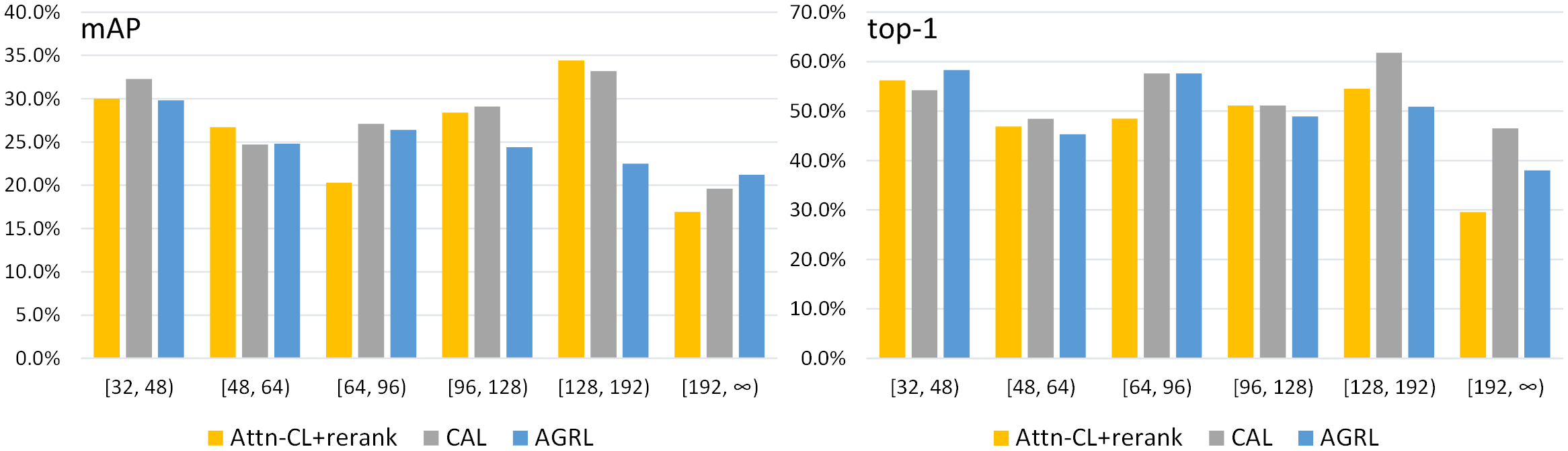}
     \caption{Video person ReID performance w.r.t. finer scales. For each experiment, the query scale is fixed to a range but the gallery can take any scale in the dataset.}
     \label{fig:fine-scales-results}
     \vspace{-10pt}
\end{figure}

\textbf{Person ReID performance w.r.t. finer scales.} To further analyze the performance to re-identify various scales of tracklets, we define $6$ finer scales based on COCO scale split~\cite{DBLP:conf/eccv/LinMBHPRDZ14}, \ie, $[32, 48)$, $[48, 64)$, $[64, 96)$, $[96, 128)$, $[128, 192)$, $[192, \infty)$. 
Then, three best performers are selected to conduct this ablation study, including Attn-CL+rerank~\cite{DBLP:conf/aaai/PathakEG20}, CAL~\cite{ccvid} and AGRL~\cite{DBLP:journals/tip/WuBLWTZ20}. 
From Figure \ref{fig:fine-scales-results}, existing methods perform similarly for majority of scales of query tracklets except the largest scale $[192, \infty)$. 
We speculate that the background noises in large bounding boxes will make negative impact on distinguishing persons, especially under the lower cameras in indoor scenes.

\subsection{Discussion}\label{sec:discussion}
Although much progress is achieved in video person ReID, the state-of-the-art methods are still unsatisfying on our \thedataset. This is because of high intra-class variances within combination of different clothes, locations and scales.
In the following, we discuss several issues of current research and potential research directions revealed by our experimental analysis.

\textbf{Change-of-clothing.} Based on our dataset, we can conclude that all state-of-the-art methods are on the rack to extract discriminative embedding for person with different outfits. 
It is necessary to use multi-modality information including pose, silhouettes, gait, faces and 3D shapes.

\textbf{Multiple challenges in complex scenes.} Current deep learning based video person ReID methods are prone to fail in our \thisdataset with multiple challenges. 
Our experiments show that, as scale changes more between the query and gallery, performance drops as well. 
There is a similar trend in significant location difference, which forces a diversity of viewpoints on each person (\eg, side view vs. nadir perspectives due to camera height and angle).

\textbf{Video person search.} In our dataset, we use ground-truth tracklets for video person ReID, which is unpractical in real life. Therefore, video person search is desired for further research.
A naive attempt is to combine the state-of-the-art multi-object tracking (MOT) methods (\eg, FairMOT~\cite{DBLP:journals/ijcv/ZhangWWZL21}, and ByteMOT~\cite{DBLP:journals/corr/abs-2110-06864}) and video-based person ReID networks. However, it is not efficient to run two separate sub-models. 
On the contrary, a unified framework with MOT and ReID can leverage joint optimization of the two modules and obtain high computation efficiency by sharing the backbone features.

\textbf{Combination with activity recognition.} To help improve ReID accuracy, we can consider vision+language models given activity labels of tracklets.  
That is, the combination of global identities and the existing MEVA activity labels~\cite{DBLP:journals/corr/abs-2012-00914} will allow for future new research into activity recognition, and multi-person interactions.
For example, it is beneficial for person ReID to capture the activity of a person exiting a vehicle, walking into a building or dropping a package and exiting a building.

%% file: conclusion.tex
\section{Conclusion}
\label{sec:conclusion}
In this work, we propose the new and challenging \thisdataset dataset for the computer vision community. Previous datasets have enabled us to push the state-of-the-art in video person ReID, but lack sufficient variety to emulate real world conditions. Our \thisdataset demonstrates the significant failure of state-of-the-art video person ReID to handle closer to real-world situations in which actors may be captured in different clothing and locations, from different viewpoints, and different scales. 
To our knowledge, it is the most varied person ReID dataset to date in terms of these factors which impact nearly all aspects of person tracking, ReID, and search. We hope the dataset can facilitate the research and development in real-life person search. Also, we release all source video and annotation tools for extending this dataset. 
Notably, by extending the MEVA dataset without compromising personally identifying information, \thedataset inherits MEVA's IRB-reviewed HSR compliance.

\ifwacvfinal
{\noindent \textbf{Acknowledgements.}}
This research is based upon work supported in part by the Office of the Director of National Intelligence (ODNI), Intelligence Advanced Research Projects Activity (IARPA), via [2017-16110300001 and 2022-21102100003]. The views and conclusions contained herein are those of the authors and should not be interpreted as necessarily representing the official policies, either expressed or implied, of ODNI, IARPA, or the U.S. Government. The U.S. Government is authorized to reproduce and distribute reprints for governmental purposes notwithstanding any copyright annotation therein.

This material is based upon work supported by the United States Air Force under Contract No. FA8650-19-C-6036. Any opinions, findings and conclusions or recommendations expressed in this material are those of the author(s) and do not necessarily reflect the views of the United States Air Force.
\fi

%% file: appendix.tex
\newpage
\clearpage

\section{Locations in MEVID}
\begin{figure*}[t]
    \centering
    \includegraphics[ width = 1.0\textwidth ]{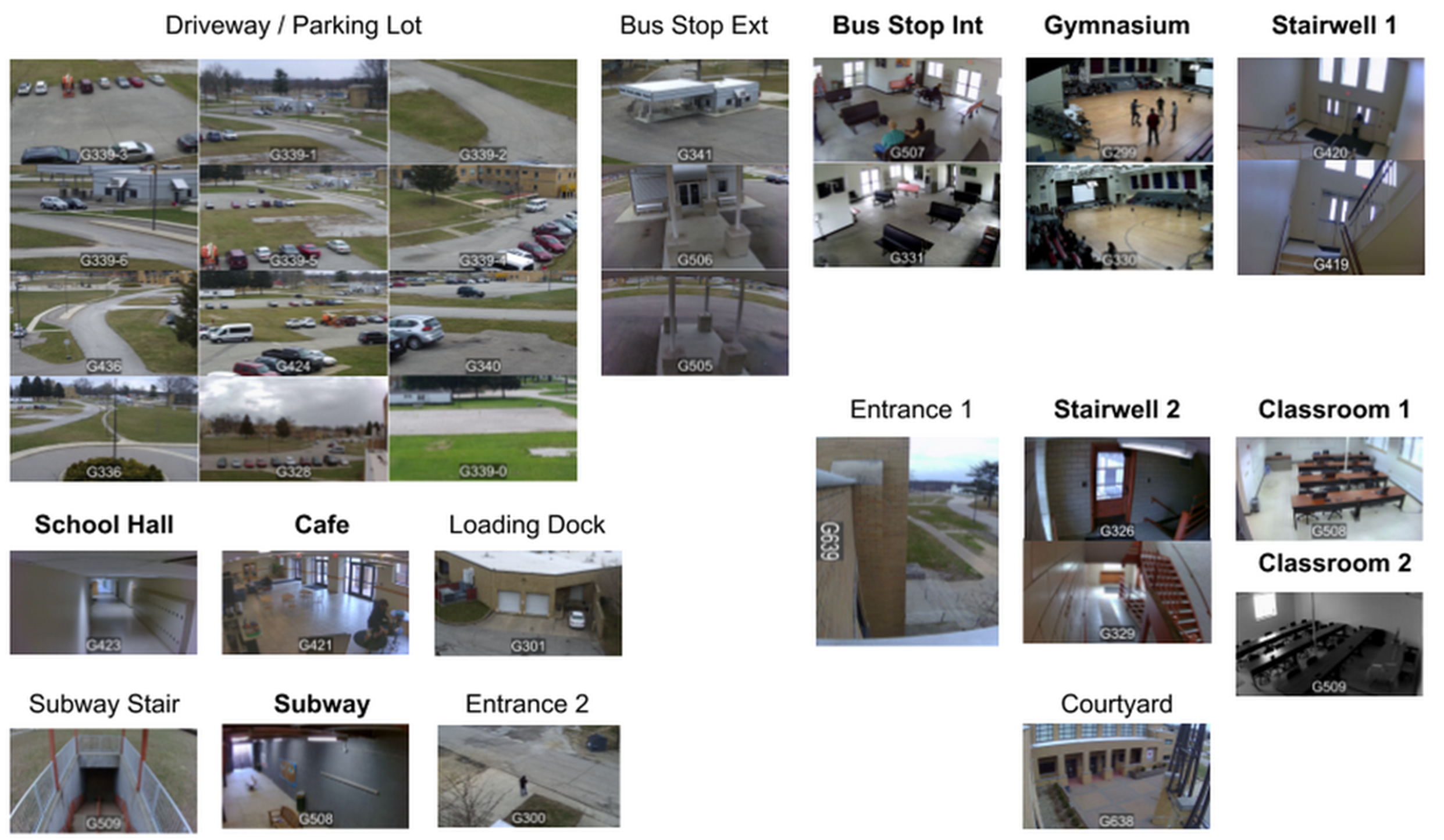}
    \caption{Screenshots of all $17$ different locations in MEVID. Note that the indoor locations are highlighted in bold font.}
    \label{fig:mevid_cameras}
\end{figure*}

As shown in Figure \ref{fig:mevid_cameras}, we display all the $17$ locations in our MEVID dataset.

\section{MEVID Persons}
\begin{figure*}[t]
    \centering

  \begin{tabular}{@{}c@{}}
    \includegraphics[width=.75\linewidth]{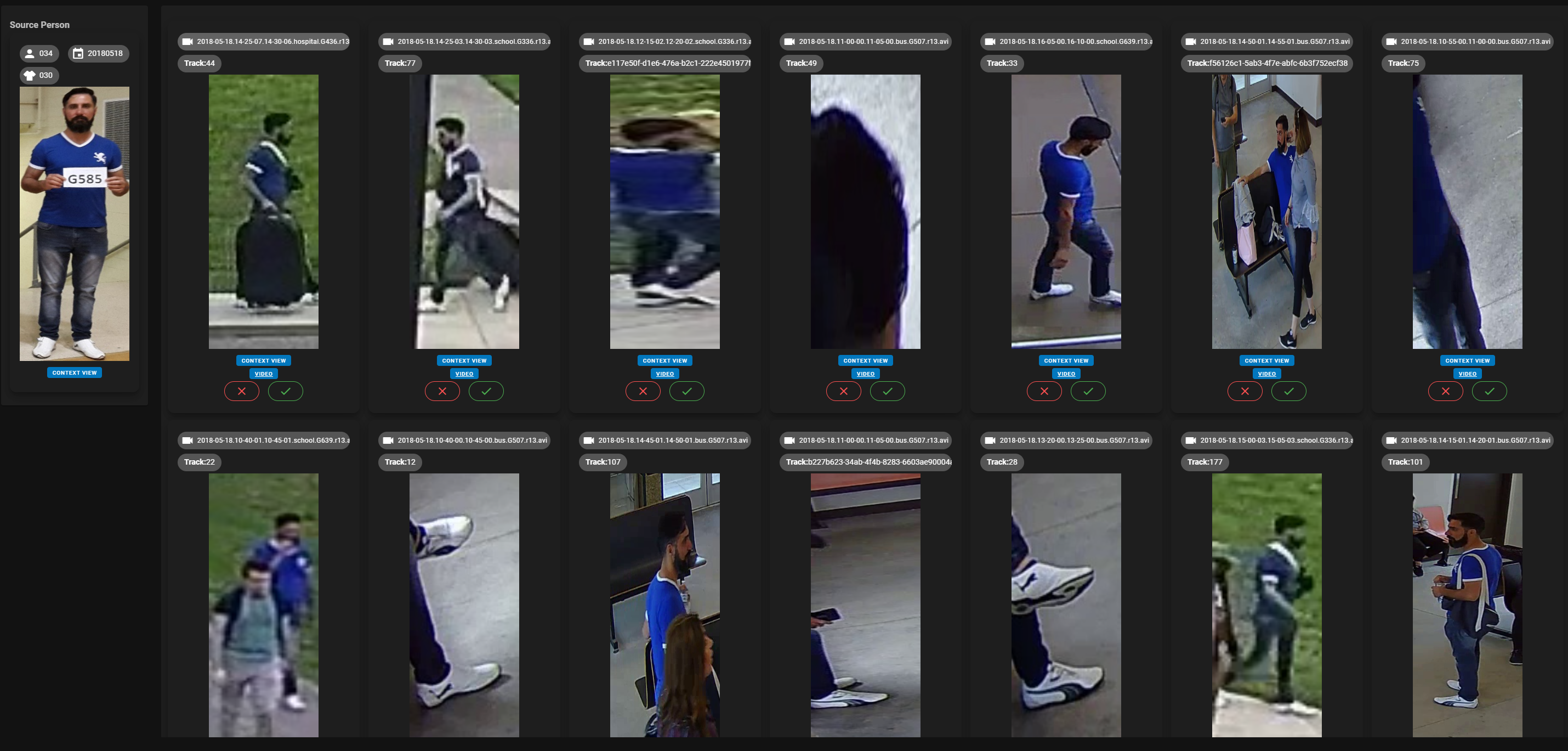} \\[\abovecaptionskip]
  \end{tabular}

  \vspace{\floatsep}

  \begin{tabular}{@{}c@{}}
    \includegraphics[width=.75\linewidth]{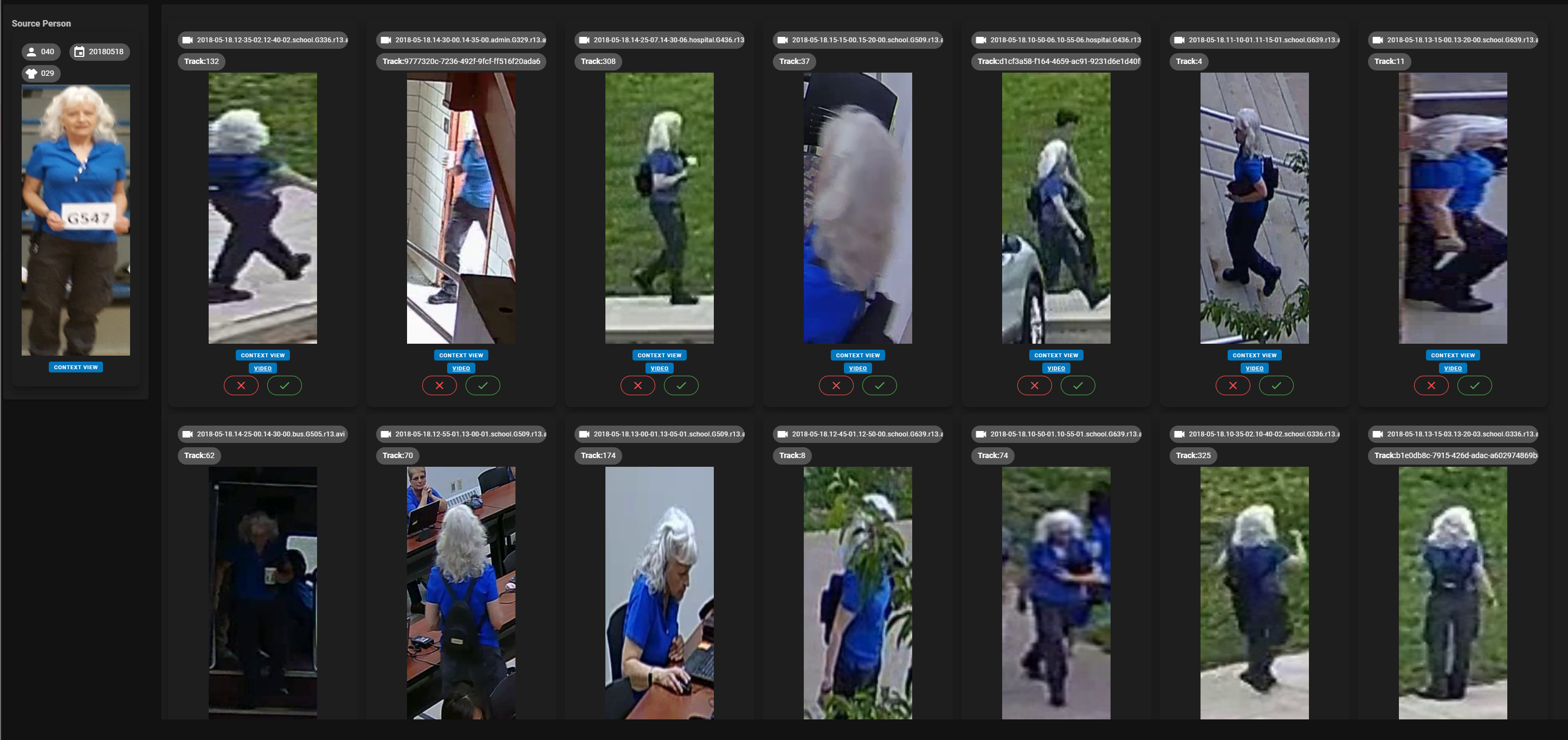} \\[\abovecaptionskip]
  \end{tabular}
  
    \vspace{\floatsep}

  \begin{tabular}{@{}c@{}}
    \includegraphics[width=.75\linewidth]{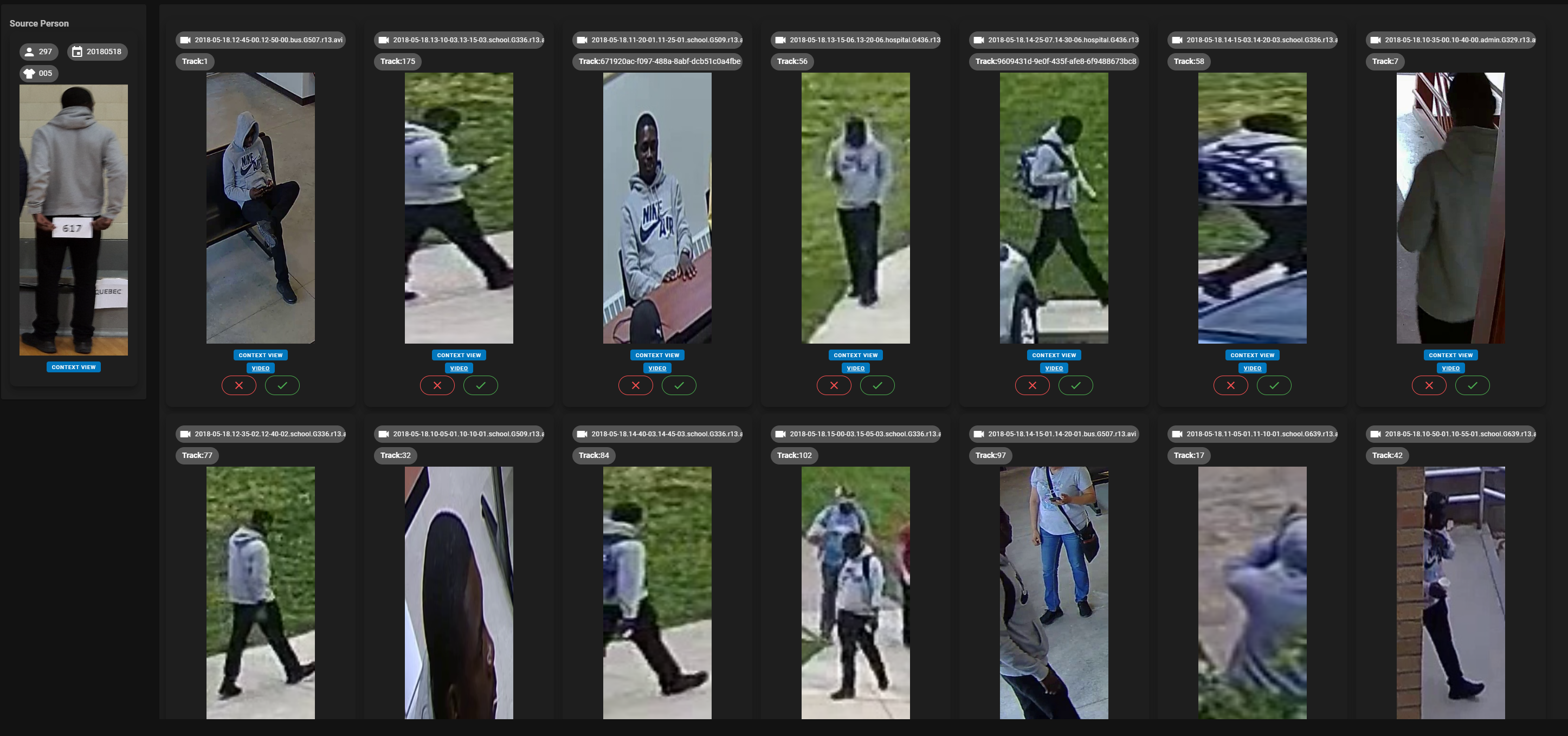} \\[\abovecaptionskip]
  \end{tabular}
  
      \caption{Several actors from the MEVID dataset and their annotated tracks.}
    \label{fig:mevid_persons}
\end{figure*}

\begin{figure*}[t]
    \centering

  \begin{tabular}{@{}c@{}}
    \includegraphics[width=\linewidth]{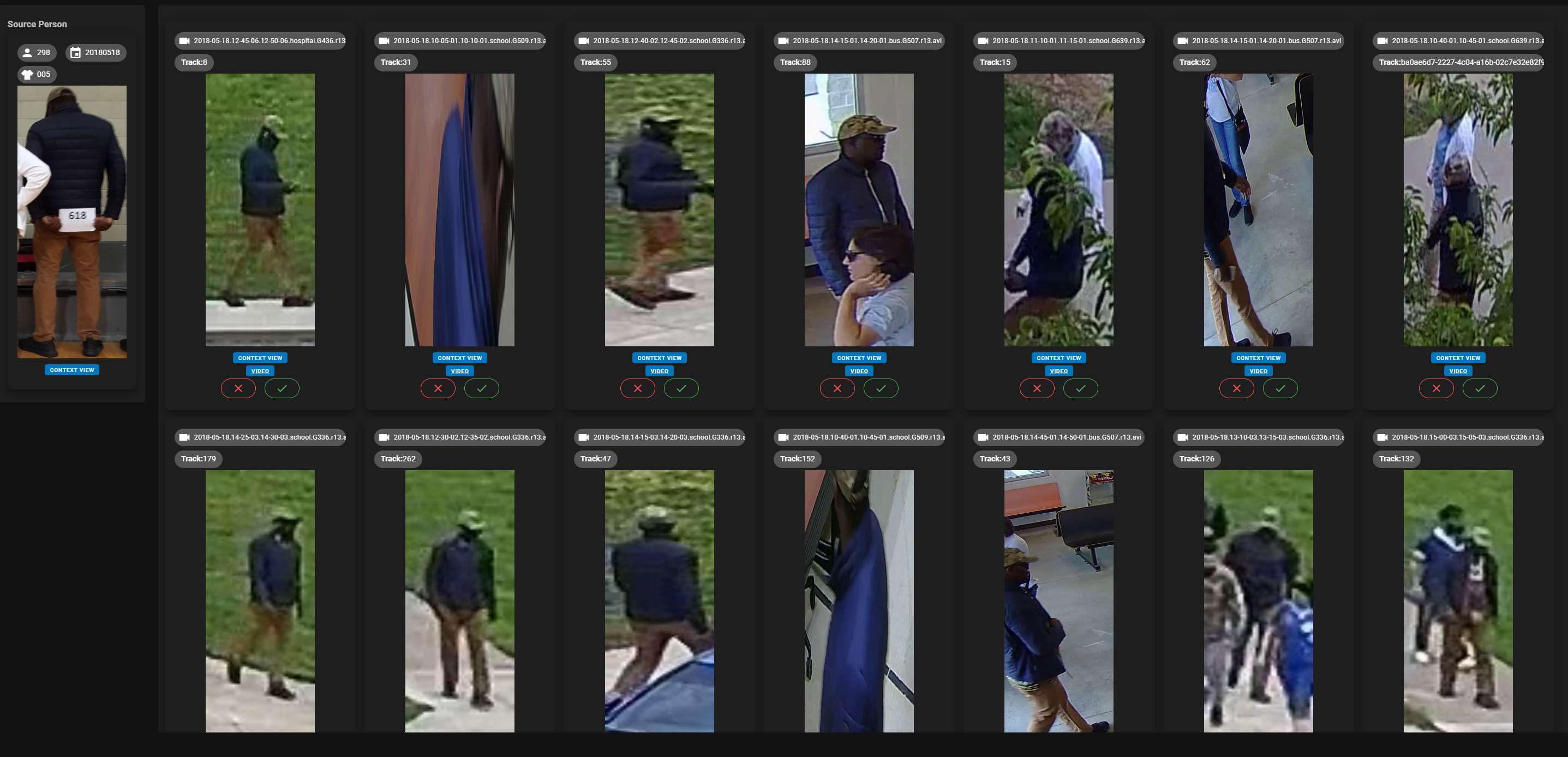} \\[\abovecaptionskip]
  \end{tabular}
  
      \vspace{\floatsep}

  \begin{tabular}{@{}c@{}}
    \includegraphics[width=.75\linewidth]{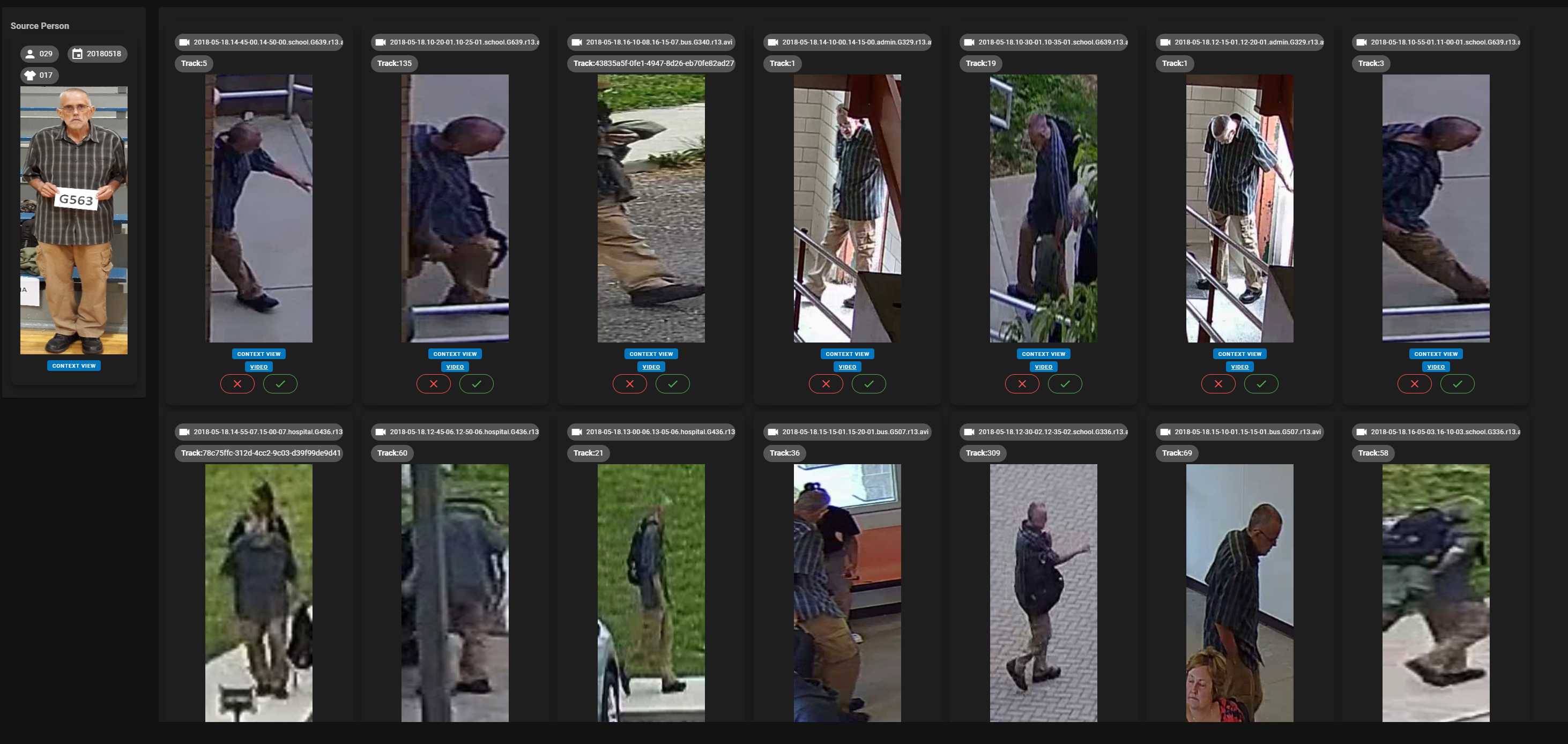} \\[\abovecaptionskip]
  \end{tabular}
  
      \vspace{\floatsep}

  \begin{tabular}{@{}c@{}}
    \includegraphics[width=.75\linewidth]{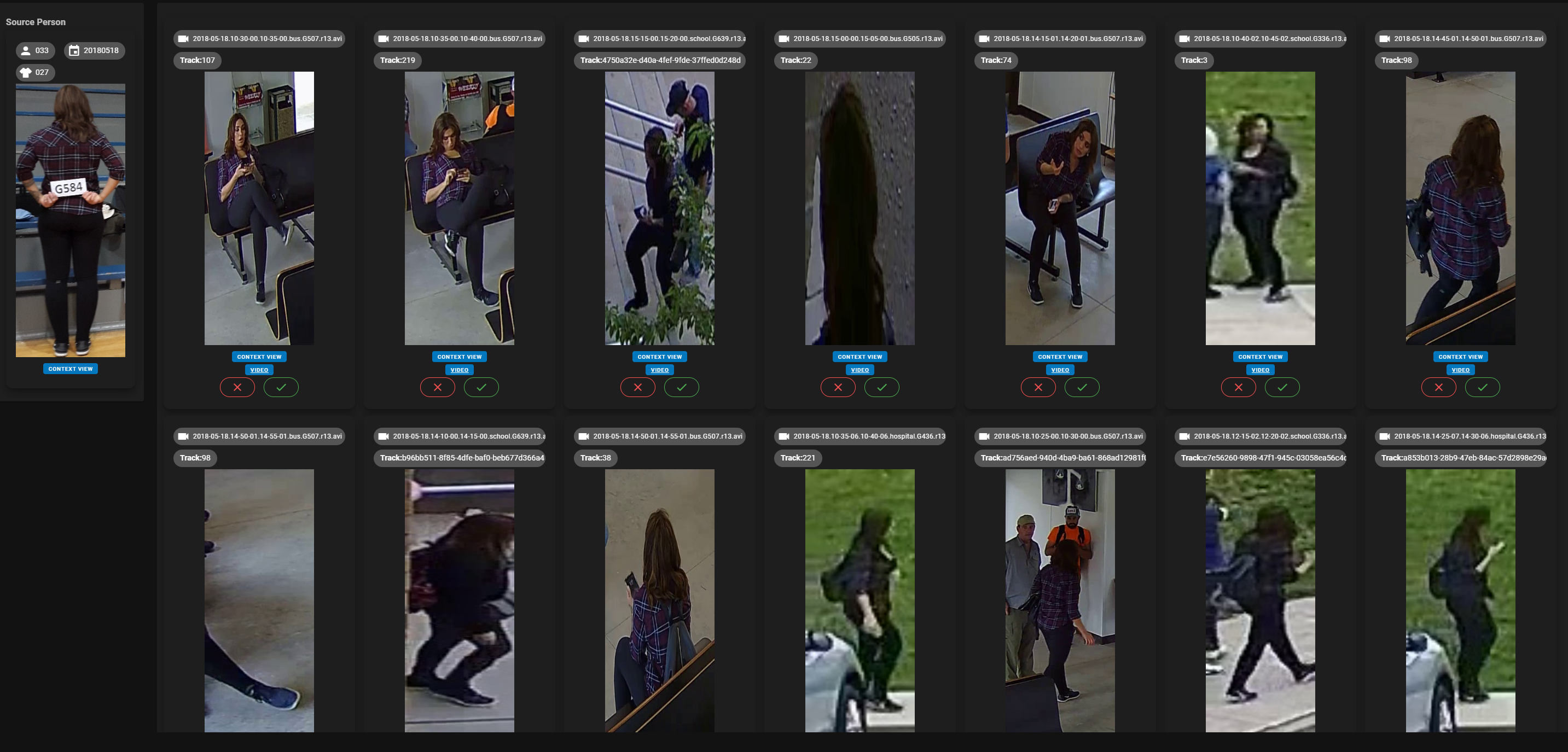} \\[\abovecaptionskip]
  \end{tabular}

    \caption{Several actors from the MEVID dataset and their annotated tracks.}
    \label{fig:mevid_persons2}
\end{figure*}

We show several of our actors and a few random tracks from each in Figure~\ref{fig:mevid_persons} and Figure~\ref{fig:mevid_persons2}. As mentioned in the main text, all persons in the MEVID dataset are from the MEVA video dataset, which was collected in a controlled closed world with hired actors whose images were collected with consent under IRB approved experimental conditions.

\section{Annotation Tool}
\begin{figure*}[t]
    \centering
    \includegraphics[ width = 1.0\textwidth ]{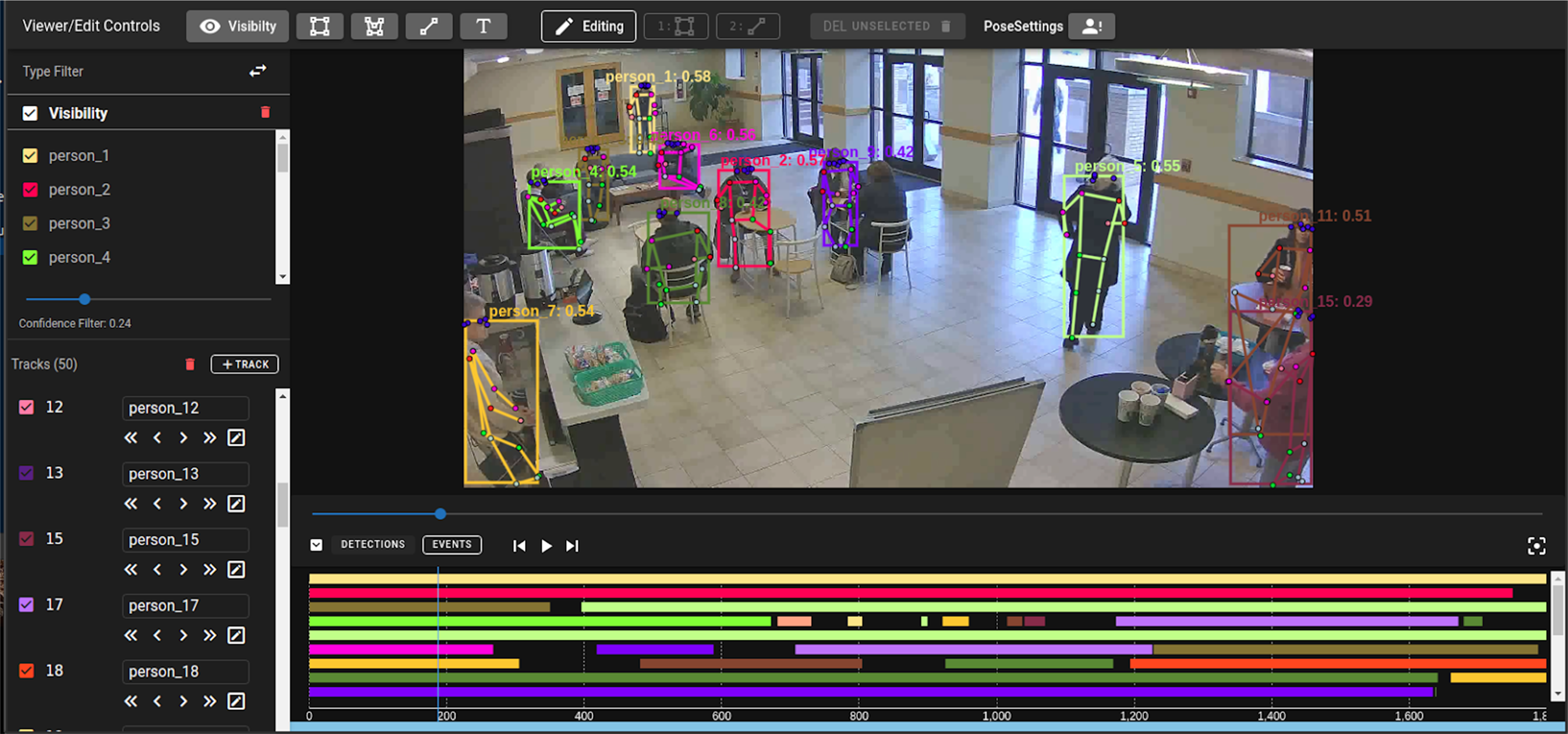}
    \caption{GUI of our proposed annotation tool for MEVID dataset.}
    \label{fig:mevid_gui}
\end{figure*}

To annotate the data, we develop the semi-automated annotation tool to ingest multiple videos, perform pose tracking, confirm and then complete person tracklet linking (see Figure \ref{fig:mevid_gui}). We attach 3 video walkthroughs of this annotation process which includes the video track cleanup, linking of tracks to checkin imagery for each person and outfit combination, and review and auditing of the links made for reid and search dataset compilation. This GUI will be release free and open source to the community alongside the dataset and baseline code.

\section{Comparison with CCVID}
As discussed in the related work section, Gu~\etal~\cite{ccvid} produced the first video change-of-clothing ReID (CCVID) dataset. However, the data is collected in fixing viewpoint, background, and scale to a relatively limited range while capturing video of actors in multiple outfits. In contrast, our dataset aims to advance state-of-the-art video person ReID algorithms in the wild, collected from a wider set of conditions of diversity in locations, actors, viewpoints, both indoor and outdoor scenes, and target scale. As presented in Table \ref{tab:ccvid-comparison}, we provide detailed comparison with CCVID~\cite{ccvid}.

\begin{table}[h]
\centering
\rowcolors{1}{gray!15}{white}
\resizebox{\columnwidth}{!}{
    \begin{tabular}{lcccccc}
    \toprule \rowcolor{LightBlue}
    \textbf{Subset}& \multicolumn{3}{c}{\textbf{CCVID}} & \multicolumn{3}{c}{\textbf{MEVID}} \\ \cmidrule(r){1-1} \cmidrule(r){2-4} \cmidrule(l){5-7}
          & \# ids   & \# tracklets & \# outfits  & \# ids   & \# tracklets & \# outfits \\
    Train   & 75  & 948  & 159  & 104  & 6338  & 485 \\
    Query  & 151  & 834  & 160  & 52  & 316  & 100   \\
    Gallery & 151  & 1074  & 252  & 54  & 1438  & 113   \\
    Total  & 226  & 2856  & 480  & 158  & 8092  & 598   \\
    \bottomrule
    \end{tabular}
    }
    \caption{Comparison with CCVID \cite{ccvid}.}
    \label{tab:ccvid-comparison}
\end{table}

